# Planning and Acting under Uncertainty: A New Model for Spoken Dialogue Systems

**Bo Zhang\*, Qingsheng Cai**
Department of Computer Science & Technology
University of Science & Technology Of China
Hefei 230027, P.R.China

**Jianfeng Mao\***
Department of Automation
Tsinghua University
Beijing 100084, P.R.China

**Baining Guo**
Microsoft Research, China
3F Sigma Center, 49 Zhichun Rd.
Beijing 100080, P.R.China

## Abstract

Uncertainty plays a central role in spoken dialogue systems. Some stochastic models like the Markov decision process (MDP) are used to model the dialogue manager. But the partially observable system state and user intentions hinder the natural representation of the dialogue state. A MDP-based system degrades quickly when uncertainty about a user's intention increases. We propose a novel dialogue model based on the partially observable Markov decision process (POMDP). We use hidden system states and user intentions as the state set, parser results and low-level information as the observation set, and domain actions and dialogue repair actions as the action set. Here, low-level information is extracted from different input modalities, including speech, keyboard, mouse, etc., using Bayesian networks. Because of the limitation of the exact algorithms, we focus on heuristic approximation algorithms and their applicability in POMDP for dialogue management. We also propose two methods for grid point selection in grid-based algorithms.

## 1 INTRODUCTION

Uncertainty plays a central role in spoken dialogue systems. The system may be uncertain about a user's intention behind a recognized utterance and also the effect of its own utterance. Although participants may tolerate a small degree of uncertainty, an excessive amount in a given context can lead to misunderstanding with different costs (Paek and Horvitz, 1999).

A dialogue manager can be formulated as a Markov decision process (MDP), where the dialogue state represents the knowledge of the system (Levin et al., 1998 & 2000; Singh et al., 2000). The MDP-based system can handle uncertainty about the effect of its own utterance, but fails to handle the uncertainty about the user's intention when it deviates from the recognized utterance in a complex environment. The reason is that the knowledge of the MDP-based system can match the user's intention only in an ideal environment.

A dialogue system should be able to carry on a conversation without the luxury of perfect speech recognition, language understanding, or precise user models (Paek & Horvitz, 1999). To handle the uncertainty emerging from the deviation of the dialogue state and the system observation, we must convert the definition of the dialogue state and find a bridge to the system observation. The partially observable Markov decision process (POMDP) framework, a model of an agent planning and acting under uncertainty, provides a systematic method of doing just that (Kaelbling et al., 1998).

Dialogue management is essentially a problem of planning and acting under uncertainty. In the POMDP framework, we define the dialogue state by a set of state variables directly representing the user's intentions and hidden system states. The observations come from different input modalities, including speech, keyboard, mouse, etc. The observation probability function serves as the bridge from states to observations.

Compared with the POMDP-based model in (Roy et al., 2000), our model adds hidden system states in addition to user intentions, which can make use of the abstract observations from multi-modality input. The construction of the state transition and observation probability function is also simplified by exploiting the use of 2TBNs (Boutilier et al., 1999). Unlike their augmented MDP approximation, we use heuristic approximation methods, which are robust and effective.

Since the number of multi-modality observations is large, using them directly will make the POMDP computationally intractable. We propose an observation

---

extraction method using Bayesian networks. A Bayesian network can combine observations from various information sources and extract abstract observations to support user barge-in and turn-taking. It reduces the number of observations without ignoring important information.

The remainder of this paper is organized as follows. In the next section, we briefly introduce the POMDP model and algorithms. In section 3, we present the dialogue manager in the form of a POMDP. The observation extraction model using Bayesian networks is described in section 4. Section 5 contains our experiments and discussions. Section 6 is devoted to the conclusion and future work.

## 2   POMDP AND ALGORITHMS

The planning problem can be defined as this: given a complete and correct model of the world dynamics and a reward structure, find an optimal way to behave (Kaelbling et al., 1998). Many planning problems can be modeled as MDPs and analyzed using the techniques of decision theory (Boutilier et al., 1999). An MDP is a model of an agent interacting synchronously with a world (Kaelbling et al., 1998). It can be specified as a tuple $<S, \mathcal{A}, T, \mathcal{R}>$, where

- $S$ is a finite set of states of the world;
- $\mathcal{A}$ is a finite set of actions;
- $T:S\times\mathcal{A}\to\Pi(S)$ is the state-transition function, given for each world state and agent action, a probability distribution over world states ( we write $T(s, a, s')$ for the probability of ending in state $s'$, given that the agent starts in state $s$ and takes the action $a$); and
- $\mathcal{R}:S\times\mathcal{A}\to R$ is the reward function, given the expected immediate reward gained by the agent for taking each action in each state (we write $R(s, a)$ for the expected reward for taking action $a$ in state $s$).

A POMDP is an MDP in which the agent is unable to observe the current state. Instead, it makes an observation based on the action and resulting state. A POMDP can be specified by extending the MDP as a tuple $<S, \mathcal{A}, T, \mathcal{R}, \Omega, O>$, where

- $S, \mathcal{A}, T,$ and $\mathcal{R}$ define an MDP;
- $\Omega$ is a finite set of observations the agent can experience in its world; and
- $O:S\times\mathcal{A}\to\Pi(\Omega)$ is the observation function, which gives, for each action and resulting state, a probability distribution over possible observations (we write $O(s', a, o)$ for the probability of making observation $o$ given that the agent took action $a$ and landed in state $s'$).

In POMDP, an agent can use a belief state to represent its knowledge of which state it may be in. A belief state $b:S\to[0,1]$ is a probability distribution over $S$. An agent uses the belief update function $\tau:B\times\Omega\times\mathcal{A}\to B$ to update its

belief state. Here $B$ is the infinite set of all the belief states, $\tau$ is defined as:

$$\tau(b,a,o)(s')=O(s',a,o)\sum_{s\in S}T(s,a,s')b(s)/Pr(o|a,b).$$

The agent is expected to gain the immediate reward

$$\rho(b,a)=\sum_{s\in S}R(s,a)b(s)$$

for taking action $a$ in belief state $b$.

A POMDP can be converted to an equivalent belief state MDP and solved by value iteration (Bellman, 1957), considering only the piecewise linear and convex (PWLC) representations of value function estimates (Sondik, 1971). Using a vector set $\Gamma_i$ to represent a PWLC function set $V_i$:

$$V_i(b)=\max_{\alpha_i\in\Gamma_i}\sum_{s\in S}b(s)\alpha_i(s),$$

value iteration becomes:

$$V_{i+1}(b)=\max_{\alpha\in A}\left\{\rho(b,a)+\gamma\sum_{o\in\Omega}\max_{\alpha_i\in\Gamma_i}\sum_{s'\in S}\left[\sum_{s\in S}T(s,a,s')O(s',a,o)b(s)\right]\alpha_i(s')\right\}$$

or

$$\alpha_{i+1}^W(s)=R(s,a)+\gamma\sum_{o\in\Omega}\sum_{s'\in S}T(s,a,s')O(s',a,o)\alpha_i^{j_o}(s')$$

to iterate in the form of the vector set directly. Here

$$W=(a,\{o_1,\alpha_i^{j_1}\},\{o_2,\alpha_i^{j_2}\},....,\{o_{|\Omega|},\alpha_i^{j_{|\Omega|}}\})$$

represents a combination of an action $a$ and a permutation of $\alpha_i$ vectors of size $|\Omega|$. In each step of the iteration, all the dominated vectors (Cassandra, 1998) are removed.

There exist many exact algorithms to solve the optimal solution for POMDP (Cassandra, 1998). The incremental pruning algorithm (Cassandra et al., 1997) is the more recent and efficient one. But it still suffers from the exponential growth of the number of the vectors used to represent the optimal value function.

Some heuristic methods approximate the optimal solution by considering only the partial vector set. We are interested in four algorithms (Hauskrecht, 2000):

- MDP approximation is the simplest way that assumes full observation of the current state. Only one vector is needed to represent the value function:

$$\alpha_{i+1}(s)=\max_{a\in A}\left[R(s,a)+\gamma\sum_{s\in S}T(s,a,s')\alpha_i(s')\right].$$

- QMDP approximation, based on the same full observation assumption, uses Q-functions as the value function for each state-action pair. So each action corresponds to one vector:

$$\alpha_{i+1}^a(s)=R(s,a)+\gamma\sum_{s'\in S}T(s,a,s')\max_{a\in A}\alpha_i^a(s').$$

- The Fast Informed Bound (FIB) method differs from the MDP and QMDP approximation in that the agent cannot know the current state of the world. Here the assumption is the full observation of future states. So we can select the best vector for every observation and every current state separately:

$$\alpha_{i+1}^a(s)=R(s,a)+\gamma\sum_{o\in\Omega}\max_{a\in A}\sum_{s\in S}T(s,a,s')O(s',a,o)\alpha_i^a(s').$$

With exact algorithms, we seek the best vector for every observation and the combination of all states.

- Grid-based approximation with linear function updates considers only the value functions of some belief states. For every belief state $b$ and action $a^1$, we update the vector set using:

$$\alpha_{i+1}^{b,a}(s) = R(s,a) + \gamma \sum_{o \in \Omega} \sum_{s' \in S} T(s,a,s')O(s',a,o)\alpha_i^{i,b,a,o}(s')\,,$$

where

$$i(b,a,o) = \arg\max_j \sum_{s \in S}\left[\sum_{s' \in S} T(s,a,s')O(s',a,o)b(s)\right]\alpha_i^j(s')\,.$$

An incremental approach (Hauskrecht, 2000) was proposed since the grid-based method is not guaranteed to converge. The idea is to keep the original vectors in the updated vector set.

Some exact methods also use a collection of linear functions for a set of belief states to represent the PWLC value function. But the exact set of belief states is difficult to initially identify. The grid-based method uses an easy-to-compute but incomplete set of belief states to approximate the optimal solution.

We consider four strategies for selecting the grid of belief state points. The first two strategies are relatively simple. The first one is the fixed-grid strategy, which chooses the extreme points of the belief state space. The second one is the random-grid strategy, which chooses a random grid at each iteration step.

We propose another two strategies based on the belief state points generated in simulation. The first one chooses the grid points randomly from the simulation points at each iteration step. We called it the random-s-grid strategy. The second one, the cluster-s-grid strategy, must cluster the simulation points first. A typical point from each cluster is chosen as the grid point. Since the belief state space is different from other multi-dimensional spaces, we also consider the entropy of the belief state in clustering:

$$\text{Dist}(b_1,b_2) = \sqrt{\text{Entropy}(b_1) * \text{Entropy}(b_2)} * \text{EDist}(b_1,b_2)\,.$$

Here $\text{EDist}(b_1, b_2)$ represents the Euclidian distance between $b_1$ and $b_2$.

# 3   POMDP FOR DIALOGUE MANAGER

Dialogue management is essentially a planning problem: the task of the dialogue manager is planning an optimal policy and acting under uncertainty. The dialogue manager, a high-level component of our spoken dialogue system, is modeled in this section using the POMDP. When we get an (near-)optimal solution of a POMDP in the form of value function, which is represented using a vector set, we can derive the (near-)optimal policy $\pi : B \rightarrow A$ from this solution. The policy will select the action that maximizes the expected reward.

---

[1] In (Hauskrecht, 2000), only one value function is used for each belief state. However, we use Q-functions for every belief state and action pair.

A simple example is used to explain the model. It also serves as the example in our experiment. It is derived from the tour guide system of the Forbidden City, the first application of the E-Partner project at Microsoft Research, China. Maggie the tour guide chooses her action according to the user's request. If she is not clear about the user's request, she can ask the user for more information using different strategies. To simplify the discussion, we only consider two kinds of requests: to visit a place, or to ask for a property of a place. Two places used in the example are a hall and a gate. The properties of these two places are their height and size.

In the following sub-sections, we propose our model for the dialogue manager as the six elements in the POMDP tuple, and compare it with the model in (Roy et al., 2000).

## 3.1   STATE

Generally, a dialogue manager must have the ability to clarify the dialogue state. It updates its state upon receiving different information from the user or the environment. Because of the inaccessible user intention and system hidden state, many dialogue managers (like MDP-based model in (Levin et al., 2000)) use the system's knowledge as the dialogue state. Usually the knowledge is gained directly from different observations including user utterances, results of database query, etc. These dialogue managers work well in the ideal environment where recognized user utterances closely reflect the user's intention. But when uncertainty increases, i.e., the environment becomes noisier or the user's task becomes more complex, the performance may degrade quickly.

In the POMDP framework, a dialogue manager can deal with the uncertainty of the exact dialogue state. So we can employ the user's intentions and other hidden system states as our dialogue states directly. The dialogue manager can update its belief state using observations extracted from the user's utterances and from other information. This makes it as easy as the MDP-based system to construct the reward function. Even more importantly, the dialogue manager is more robust in handling the uncertainty emerging from the deviation between user's utterances and intentions.

We use a factored representation of our dialogue state space (Boutilier et al, 1999). In our example, dialogue states, which are also POMDP states, consist of two independent parts: user's intentions and hidden system states. We use three state variables to represent the user's intention. They are the request type (visit or ask), the place (gate or hall), and the property (height or size). The hidden system states include normal, silent, error (noisy), error (silent), and overhead. Altogether there are 40 states, among which 10 state pairs are equivalent pairs, since the property variable is useless when the value of the type is equal to "visit".

## 3.2 ACTION

We divide actions in our dialogue system into two classes: actions for satisfying the user's request, and actions for gathering more information from the user to clarify the user's intention. Actions belonging to the first class are domain actions and are usually simple, and actions in the second class are known as repair actions. Selecting appropriate repair actions is very important to the success of a spoken dialogue system working in a complex environment.

In our example, we define two actions in the first class: answering the user's question and changing the place at the user's request. Repair actions include asking for the user to repeat the statement, asking for the user's intention (type, place or property), declaring the user's intention, ignoring the user, and trouble-shooting (executed when the dialogue manager believes the speech recognizer does not work properly, i.e. microphone fails to work). The total number of actions is 18.

## 3.3 STATE TRANSITION FUNCTION

We have two assumptions on the state transition function. First, we assume that the user's intention does not change until the request is processed. The repair actions do not change the user's intention. The second assumption is that only the troubleshooting action is related to the hidden system state. Other actions do not affect the transitions among the hidden system states. These two assumptions greatly simplify the design of the state transition function.

Since we use a factored representation of the state space, we can use a two-stage temporal Bayesian network (2TBN) (Boutilier et al., 1999) to specify the state transition function for every action. Some actions of the same type can share the same 2TBN.

In our example, we have only three simple 2TBNs for the 18 actions. To demonstrate the ability of handling the tremendous uncertainty, we design the hidden state transition function by intentionally increasing the possibility of falling into abnormal states. This model is used in the simulation and our system turns out to be very robust. In real world applications, the state transition function must reflect the properties of the system and the environment, i.e. both the hardware and software of the speech recognizer.

## 3.4 OBSERVATION

Observations come from recognized user's utterances and other low-level information contained in the speech recognition result, parser result, keyboard and mouse input, etc. Since the structures of these observations are different from each other, to combine them in the POMDP framework, we must extract some abstract observations from various information sources. Simply ignoring some useful information like confidence in the speech recognition result is not wise.

Like the Quartet architecture (Paek & Horvitz, 2000), we use a channel level and a signal level as the lower levels of the spoken dialogue system. Bayesian networks are used to infer the status of each level from the low-level information. In the channel level, the system can be in "Channel" or "No channel" status; in the signal level, it can be in "Signal" or "No signal" status. So we have four possible observations now. When the system is in "Channel" and "Signal" status, we divide this observation into more detailed observations, which come from the parser result of the recognized user's utterance.

In our example, 22 observations come directly from the user's utterances, including affirmative answers, negative answers, and (incomplete) user requests, which may be any meaningful combination of the type, place and/or property of the request. So the POMDP model includes 25 observations.

## 3.5 OBSERVATION PROBABILITY FUNCTION

The most complex part of the POMDP model for the spoken dialogue system is the observation probability function. The same action may lead to different observations even in the same state. One reason is that the speech recognizer is far from perfect. To make things worse, different users, or even the same user at different times, tend to provide different answers for the same question. So the construction of the observation probability function requires deep insight of the speech recognizer and a good user model.

We also use 2TBNs to construct the observation probability function. Eleven 2TBNs are used in our example, among which six are for declaring user intention actions and three are for asking user intention actions. 2TBNs in the same group are very similar. All of them are handcrafted, depending a lot on the experience of the developer.

## 3.6 REWARD FUNCTION

The reward function is relatively simple. We need only specify the rewards of a particular action executed in a particular state, i.e. a positive reward when the answer matches the user's request, or a negative reward (cost) if a mismatch occurs. Repair actions are also associated with negative rewards.

In our example, 11 different rewards are specified. These rewards belong to two classes: 1) for repair actions:

asking the user to repeat, asking for the user's intention, declaring the user's intention, ignoring (right/wrong), and troubleshooting (right/wrong); and 2) for domain actions: wrong type, right type without a right place or property, right type with a right place or property, and totally right.

### 3.7 COMPARISON WITH OTHER MODEL

In this section, we compare our model with the model used in (Roy et al., 2000). In addition to user intentions they used to construct the POMDP state space, we also consider hidden system states, which are useful in complex environments since they make use of observations from low-level information. Our factored representation differs from their flat state space. It simplifies the construction of the state transition function and observation probability function by exploiting the use of 2TBNs. It is also much easier to adjust the parameters.

The definition of the observations is also different. Besides utterances that can reflect the user's (partial) intention, we also consider other observations inferred from the low-level information of the speech recognizer, robust parser and other input modalities. It can improve the robustness of the system and make it easier to include more input modalities like visual input from a video camera.

Roy et al. use an augmented MDP to approximate the original POMDP. They replace the belief state with a pair consisting of the most likely state and the entropy of the belief state. This approach can be applied to only some POMDPs. In the POMDP for our example, some states are equivalent. So the entropy cannot fully reflect the degree of uncertainty of the current belief state. We use several approximation algorithms to solve the POMDP. Among them, the grid-based algorithm turns out to be effective and adaptive.

## 4    BAYESIAN NETWORKS FOR OBSERVATION EXTRACTION

Low-level observations extracted from different input modalities are very important for handling the uncertainty in a spoken dialogue system. We use Bayesian networks to extract these low-level observations in the channel level and signal level (Paek & Horvitz, 2000).

The existence of a channel for communication reflects the channel level status, and the existence of a signal reflects the signal level status. The status of the channel level is primary inferred from the user's focus, and the status of the signal level is relevant to the confidence of the speech recognition result, the parser result, etc.

We want to know the status of these two levels at two critical time points. The first one is for user barge-in detection and the second one is for general turn-taking

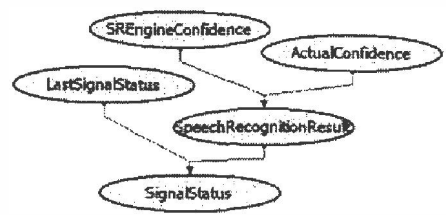

Figure 1: Bayesian Network for User Barge-in Detection

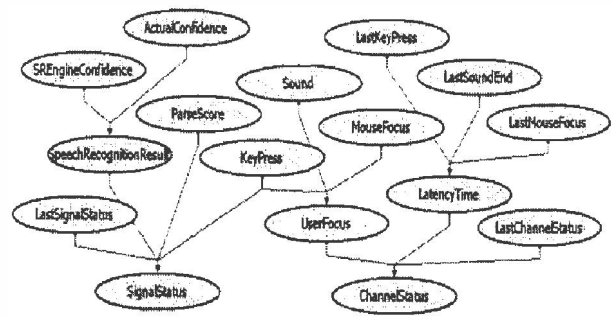

Figure 2: Bayesian Network for Turn-taking

between the system and the user.

Since we have only limited input modalities--speech, keyboard and mouse--the Bayesian network for the channel level is quite simple. It can be extended when we want to add more input modalities like a video camera to detect eye gaze. Our current focus is the more complex signal level.

To support user barge-in in a noisy environment, we must detect the user's voice before we get the recognition result. Upon receiving the sound start event, the confidence of the following three hypotheses are checked, from which the status of the signal level can be inferred (Figure 1). In Microsoft Speech SDK we use, the confidence consists of two parts: ActuralConfidence (AC) is a binary number and SREngineConfidence (EC) is a real number.

Upon receiving the recognition result, we check the confidence of its elements, its parser score, etc. A slightly different Bayesian network (Figure 2) is used to infer the status of the signal and channel level for turn-taking.

One advantage of the Bayesian network is that the result of status is a probability distribution instead of an exact state. We can adjust the threshold to tune the system. Another advantage is that our system is easy to extend, e.g. we need only add a node to the Bayesian network for the channel level and change some probability distributions if we want to add eye gaze information.

## 5    EXPERIMENTS AND DISCUSSIONS

Our experiments include two parts: the real world experiment of observation extraction using a Bayesian network in the signal level, and the simulated experiment of POMDP-based dialogue management.

Table 1: Observation Extraction Results (PS = parser score; Signal = probability of having signal; Recognized utterance)

| | 1 | 2 | 3 | Ave | 1 | 2 | 3 | 4 | Ave |
|---|---|---|---|---|---|---|---|---|---|
| AC | 0 | 0 | 0 | 0 | 0 | 0 | 1 | 1 | 0.5 |
| EC | 2030 | 705 | 2088 | 1608 | 2088 | 6092 | 1969 | 1473 | 10652 |

(a) Overheard: PS=0; Signal=0.14/0.238; "a free show half"

| | 1 | 2 | 3 | Ave | 1 | 2 | Ave |
|---|---|---|---|---|---|---|---|
| AC | 0 | 0 | 0 | 0 | 0 | 0 | 0 |
| EC | 771 | 771 | 771 | 771 | 771 | -6018 | -2624 |

(b) Noise: PS=0; Signal = 0.14/0.081; "sixty two"

| | 1 | 2 | 3 | Ave | 1 | 2 | 3 | 4 | 5 | 6 | 7 | Ave |
|---|---|---|---|---|---|---|---|---|---|---|---|---|
| AC | 0 | 1 | 1 | 0.67 | 1 | 0 | 0 | 1 | 1 | 1 | 1 | 0.71 |
| EC | 2472 | 13959 | 13959 | 10130 | 13959 | 6541 | -11052 | 46317 | 39444 | 18937 | 16548 | 18671 |

(c) Request 1(I want to go to the hall): PS=714; Signal=0.736/0.874; "I want to go to the whole"

| | 1 | 2 | 3 | Ave | 1 | 2 | 3 | 4 | 5 | 6 | 7 | 8 | 9 | 10 | Ave |
|---|---|---|---|---|---|---|---|---|---|---|---|---|---|---|---|
| AC | 1 | 1 | 1 | 1 | 1 | 1 | 1 | 1 | 1 | 1 | 1 | 1 | 1 | 1 | 1 |
| EC | 16787 | 19501 | 18058 | 18115 | 18058 | 39978 | 21437 | 41486 | 44326 | 28752 | 32239 | 33211 | 26702 | 17259 | 30345 |

(d) Request 2(Can you tell me the size of the gate): PS=600; Signal=0.859/0.874;"Can you tell me the size of the day to"

Table 2: Comparison of Four Approximation Methods (random-s-grid strategy used in grid-based method)

| | Value Function | | | Policy Quality | | Reaction Time | |
|---|---|---|---|---|---|---|---|
| | Time | Size | Average | DR | LA | DR | LA |
| MDP | 1 | 1 | 139.98 | N/A | 5454 | N/A | 489 |
| QMDP | 1 | 18 | 125.54 | -1985 | 19049 | 1 | 553 |
| FIB | 332 | 18 | 65.33 | -1232 | 25025 | 2 | 603 |
| Grid | 1380 | 290 | 19.36 | 26533 | 28986 | 3 | 952 |

## 5.1 Observation Extraction

We present the result of signal level observation extraction in four situations, including overheard, noise (like cough), and two user requests (Table 1).

We can identify user barge-in from the overheard and noise by checking the status of the signal level at the first time point (the first half of the table). At the second time point (second half), the final observation of this turn is also successfully extracted, and is fed into the dialogue manager.

## 5.2 Dialogue Management

The POMDP for our example has 40 states, 18 actions, and 253 observations. We use 0.9 as the discount factor, a fixed initial belief state, in which the probabilities of 8 user intentions satisfy the uniform distribution.

We are not surprised that all of the exact algorithms including the incremental pruning algorithm (Cassandra et al., 1997) cannot solve this POMDP, since the numbers of both actions and observations are too large for POMDP problems.

We compare four methods: MDP, QMDP, FIB, and grid-based approximation. Both the standard and incremental approaches, all of the four grid point selection strategies are used for the grid-based method. We stop the execution of the grid-based method after 30 epochs, while for other methods we can get the converged solution.

We evaluate the solution from three perspectives: running time, quality of the value function, and quality of the policy derived from the value function. The criterion for the quality of the value function is the average value in 10000 random belief points. Since the grid-based method bounds the minimum of the optimal value function, a bigger value means higher quality; for the other three methods, since they bound the maximum of the optimal value function, a smaller value means higher quality. We use direct and look-ahead methods (Hauskrecht, 2000) to get the policy from the value function. The direct method (DR) chooses the action associated with the best vector[2] while the look-ahead method (LA) extracts the policy from the vector set via a greedy one-step look-ahead:

$$\pi(b) = \arg\max_{a \in A}\left[ \rho(b,a) + \gamma \sum_{o \in \Omega} Pr(o \mid a, b) V_t(\tau(b,o,a)) \right].$$

The quality of the policy is reflected by the total rewards gained in the 10000-step simulation. The simulation is performed in a simulator, which can execute the policy in a simulated world based on the POMDP model.

To examine the adaptability of these approximation algorithms in different models for spoken dialogue system, we change the settings of POMDP in two ways. One is to add even more uncertainty to the system by changing both the state transition function and observation probability function, and increasing the presence of both abnormal states and false observations. The other is changing the reward structure by reducing the cost of wrong actions. So we have four models: standard, lower-cost, noisy, and noisy-lower-cost.

We present only a portion of the experiment results because of space limitations. From table 2, we can see that the grid-based method results in the best policy, although its speed is also the slowest. Actually, since state transition in our spoken dialogue system is not very frequent, the primary task is to clarify the dialogue state through appropriate actions and following observations. The assumption of a fully or partially observable dialogue state is not appropriate here.

---

[2] The direct method cannot be applied to the value function with only one vector in the MDP approximation method.

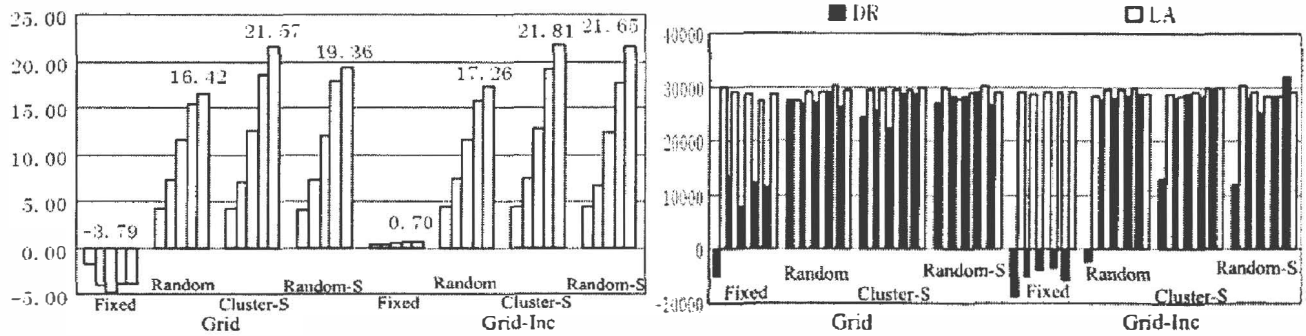

Figure 3: POMDP Solution (grid-based method, epoch 2,5,10,20,30) – Average Value and Policy Quality

Table 3: Rewards Gained in Simulation of Different Policies

| | Reward | MDP LA | QMDP DR | QMDP LA | FIB DR | FIB LA | Grid DR | Grid LA |
|---|---|---|---|---|---|---|---|---|
| Asking for repeat | -4 | 0 | 0 | 824 | 0 | 0 | 1 | 1011 |
| Ignoring (wrong) | -3 | 359 | 98 | 55 | 141 | 103 | 13 | 13 |
| Asking for user's intention | -2 | 0 | 0 | 850 | 0 | 173 | 2982 | 2534 |
| Declaring user's intention | -1 | 4454 | 7616 | 3125 | 7584 | 5045 | 1690 | 1068 |
| Ignoring (right) | 0 | 3599 | 1331 | 1753 | 1488 | 1196 | 669 | 707 |
| Trouble-shooting or Action (wrong) | -20 | 62 | 201 | 96 | 102 | 125 | 274 | 208 |
| Action (right type, no right param) | -15 | 108 | 21 | 122 | 22 | 90 | 256 | 106 |
| Action (right type, has right param) | -10 | 35 | 9 | 51 | 9 | 53 | 84 | 89 |
| Domain Action (right) | 10 | 1383 | 419 | 3124 | 391 | 2938 | 3709 | 3952 |
| Trouble-shooting (right) | 20 | 0 | 305 | 0 | 263 | 277 | 322 | 312 |
| Total Rewards | | 5454 | -1985 | 19049 | -1232 | 25025 | 26533 | 28986 |

Table 4: Observations gained in different models

| Models | No info | Yes/No | Partial | Full |
|---|---|---|---|---|
| Standard | 4371 | 1062 | 1444 | 3123 |
| Lower-cost | 4282 | 776 | 1784 | 3158 |
| Noisy | 6542 | 1282 | 1102 | 1074 |
| Noisy-lower-cost | 6577 | 611 | 1280 | 1532 |

From the simulation result (Table 3), we discover that the MDP look-ahead policy, QMDP and FIB direct policies never select effective information gathering actions with high cost like asking for a repeat or asking for the user's intention. They only choose other less effective actions with lower cost. The QMDP and FIB look-ahead policies are more effective, while both of the grid-based policies are most effective in using the information gathering actions.

Because of the intentionally increased uncertainty in our model, the dialogue manager can get less useful observations than in normal situations (Table 4). But the grid-based approximation algorithm performances quite well even if it may get false observations among such limited observations.

We can compare the different grid point selection strategies in a grid-based method (Figure 3). Obviously, the simplest fixed-grid strategy is the worst one. From the perspective of value function quality, the cluster-s-grid strategy is the best; the random-s-grid and random-grid strategies are also good. But from the perspective of policy quality, all these three strategies are almost the same. After enough epochs, the direct policy works as well as the look-ahead policy with a much shorter reaction time.

The incremental approach is a little bit better than standard approach in value function quality. But it yields similar policy quality and a longer reaction time.

The experiments in other models similarly reveal the advantages the grid-based method has over other methods (Table 5). It proves that the grid-based algorithm and our grid point selection strategies are robust. They can be applied to different models with similar performance.

# 6　CONCLUSION

In this paper, we propose a novel POMDP-based model for a spoken dialogue system. In this model, the dialogue state represents the user's intentions and hidden system states directly; the observations come from different input modalities; and the observation probability function serves as the bridge from states to observations. Abstract observations are extracted from different input modalities using Bayesian networks. Dialogue strategies are derived from the near-optimal value function of POMDP, solved by approximation algorithms. Both the real world experiment of observation extraction and the simulation of dialogue management provide positive evidence of its robustness and effectiveness in complex environments.

Table 5: Comparison of the Four Models

|  | Average Value | | | | Policy Quality (DR) | | | Policy Quality (LA) | | | |
|---|---|---|---|---|---|---|---|---|---|---|---|
|  | MDP | QMDP | FIB | Grid | QMDP | FIB | Grid | MDP | QMDP | FIB | Grid |
| Standard | 140 | 125.5 | 65.33 | 19.36 | -1985 | -1232 | 26533 | 5454 | 19049 | 25025 | 28986 |
| Lower-cost | 140 | 128.1 | 66.94 | 27.5 | 2333 | -288 | 32775 | 10419 | 19488 | 29199 | 31897 |
| Noisy | 140 | 125.7 | 61.69 | 6.051 | -1796 | -92 | 10393 | -1541 | 2693 | 10856 | 12407 |
| Noisy-lower-cost | 140 | 128.2 | 63.53 | 13.16 | 3241 | 1121 | 15482 | 233 | 4625 | 11934 | 15054 |

It is an emerging direction to use stochastic models like POMDP for dialogue management. We are considering the following open questions worthy of further investigation. A real world experiment of dialogue management is our immediate focus to fully examine the model, while the simulation experiment with a scaled-up example is also useful for qualifying the appropriate approximation algorithms. Some algorithms based on the factored state space (Boutilier & Poole, 1996) are natural candidates since we already have a factored state space but do not exploit it in the approximating. Another important problem is the construction of the model. The correctness of the model is highly reliant on the handcrafted POMDP and Bayesian networks. The combination of effective user study and some machine learning techniques are useful for attacking this problem (Singh et al., 2000).

### Acknowledgments

Our thanks to the Adaptive Systems & Interaction Group of Microsoft Research for the belief network authoring tool MSBNX and for prompt answers to our questions about it. Both our speech SDK and robust parser come from the Speech Technology Group of Microsoft Research. We also thank Tony Cassandra for making his POMDP code publicly available.